\documentclass{article}

\usepackage{arxiv}

\usepackage[utf8]{inputenc} 
\usepackage[T1]{fontenc}    
\usepackage{hyperref}       
\usepackage{url}            
\usepackage{booktabs}       
\usepackage{amsfonts}       
\usepackage{nicefrac}       
\usepackage{microtype}      
\usepackage{lipsum}		
\usepackage{graphicx}
\usepackage{subcaption}
\usepackage{natbib}
\usepackage{doi}
 \usepackage{amsmath}

\title{Deep Multimodal Wearable Sensor Fusion for Detection of Body-Focused Repetitive Behaviors}

\date{} 					

\author{
\normalfont\large
\href{https://orcid.org/0009-0002-7066-5203}{%
\includegraphics[scale=0.06]{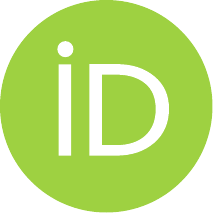}\hspace{1mm}Samaneh Rezaeimanesh}%
\thanks{These authors contributed equally to this work.}
\hspace{0.5em}
\href{https://orcid.org/0009-0002-3315-8325}{%
\includegraphics[scale=0.06]{orcid.pdf}\hspace{1mm}Mohsen Behradfar}%
\footnotemark[1]
\hspace{0.5em}
\href{https://orcid.org/0000-0001-6939-4157}{%
\includegraphics[scale=0.06]{orcid.pdf}\hspace{1mm}Mohammad Fili}
\hspace{0.5em}
\href{https://orcid.org/0000-0001-8392-8442}{%
\includegraphics[scale=0.06]{orcid.pdf}\hspace{1mm}Guiping Hu}
\\[2ex]
Department of Systems Engineering and Operations Research\\
George Mason University, Fairfax, VA 22030, USA
}

\renewcommand{\shorttitle}{Deep Multimodal Wearable Sensor Fusion for Detection of Body-Focused Repetitive Behaviors}

\hypersetup{
pdftitle={A template for the arxiv style},
pdfsubject={q-bio.NC, q-bio.QM},
pdfauthor={David S.~Hippocampus, Elias D.~Striatum},
pdfkeywords={First keyword, Second keyword, More},
}

\begin{document}
\maketitle

\begingroup
\renewcommand\thefootnote{}
\footnotetext{This work has been submitted to the IEEE for possible publication. Copyright may be transferred without notice, after which this version may no longer be accessible.}
\addtocounter{footnote}{-1}
\endgroup

\begin{abstract}
Body-focused repetitive behaviors, such as hair pulling and skin picking, are compulsive motor actions commonly associated with obsessive-compulsive and anxiety disorders. Their early, objective detection remains difficult because the movements are subtle and overlap with ordinary, non-pathological gestures. We developed and evaluated a multimodal deep learning framework to detect and classify these behaviors from wrist-worn sensor data. The data, collected by the Child Mind Institute using the Helios wrist-worn device, combine inertial measurement units, thermopile sensors, and time-of-flight sensors, capturing kinematic, thermal, and proximity information. The framework combined a convolutional neural network with a gated recurrent unit, alongside modality-specific autoencoders and a late-fusion classifier, to exploit temporal and spatial dynamics. It achieved an F1 score of 0.985 and an area under the receiver operating characteristic curve of 0.997 for binary detection, distinguishing these behaviors from other activities, and a macro-averaged F1 score of 0.700 with an area under the curve of 0.963 across a nine-class scheme that distinguished each individual behavior from a single grouped Non-Target class, improving over single-modality baselines. Post-hoc interpretability based on Shapley additive explanations showed that the time-of-flight and inertial modalities dominated discriminative power by capturing spatial proximity and dynamic movement, while hierarchical clustering indicated that misclassifications were driven primarily by the anatomical region of the gesture. These findings demonstrate that multimodal sensor fusion enables accurate, objective, and continuous behavioral monitoring. This work establishes a foundation for real-time, wearable-assisted mental health diagnostics and personalized interventions in biomedical research and clinical care.
\end{abstract}


\section{Introduction}
\label{sec:introduction}
Body-focused repetitive behaviors (BFRBs), such as hair pulling (trichotillomania) and skin picking (excoriation disorder), are compulsive, self-directed motor habits grouped among obsessive-compulsive and related disorders. As clinically recognized conditions they are relatively common: a large US survey identified current trichotillomania in 1.7\% and skin picking disorder in 2.1\% of adults, both with high psychiatric comorbidity \cite{1, 2}. Under broader definitions, the footprint grows, with up to 24\% of a community sample meeting criteria for a clinically significant BFRB disorder and 97.1\% reporting at least one BFRB in their lifetime \cite{3}. Comorbid anxiety and obsessive-compulsive disorder (OCD) are frequent \cite{8}, and stress-related markers such as heart rate variability distinguish OCD patients and predict treatment response, positioning BFRBs as indicators of emotional dysregulation \cite{9}. Clinically, these behaviors cause tangible harm, tissue damage, scarring, and infection, alongside psychosocial consequences including diminished self-esteem and social avoidance \cite{4, 5}. Yet their assessment remains difficult: self-report questionnaires and clinical interviews are limited by subjectivity, recall bias, and the episodic, frequently unnoticed nature of the behaviors, which delays timely diagnosis and intervention \cite{6}.

Wearable sensor technologies offer a promising path toward objective, continuous monitoring, using physiological and kinematic signals to detect behaviors in real time and support personalized mental health care \cite{7}. Such systems have already been applied across mental-health contexts, ranging from physiological stress detection \cite{13} to broader behavioral pattern recognition. However, BFRBs pose a harder problem than conventional human activity recognition (HAR) tasks such as gait analysis or exercise tracking: the target gestures are subtle, context-specific, and closely resemble benign everyday movements, so they must be separated from a background of similar non-pathological actions \cite{10}. Reliable detection therefore depends less on recognizing coarse activity than on capturing the fine-grained spatial and contact cues that set a pathological gesture apart.

Meeting this requirement exposes the limits of motion sensing alone. Traditional inertial measurement units (IMUs), comprising accelerometers and gyroscopes, capture how the wrist moves but not where the hand is relative to the body, and thus lack the spatial context needed to resolve the fine-grained hand-to-body interactions characteristic of trichotillomania or excoriation disorder \cite{11}. Complementary modalities close this gap: thermopile (THM) sensors measure infrared temperature and reveal skin contact or close proximity, while time-of-flight (ToF) sensors provide precise distance measurements that localize hand-to-face and hand-to-body movements across varied postures \cite{11, 19}. Prior work confirms the value of this fusion, adding thermal sensing improved wrist-worn tracking of hand position relative to the head beyond what IMU and proximity data achieved alone \cite{11}, and, more broadly, integrating auxiliary modalities with IMUs consistently enhances gesture-detection accuracy over unimodal systems \cite{12, 27}.

Translating these multimodal streams into reliable predictions calls for models that jointly exploit their temporal and spatial structure. Systematic reviews emphasize that fusing data across sensing modalities is critical for detecting complex behavioral and mental-health patterns, providing multi-dimensional context that single-sensor systems cannot \cite{14}. Combining IMU data with physiological or environmental inputs has yielded granular insight into daily activities, fatigue, and cognitive states, surpassing unimodal approaches \cite{15, 16}. Deep learning is particularly well suited to this task: hybrid convolutional neural network-gated recurrent unit (CNN-GRU) architectures capitalize on both spatial features and temporal sequences to classify intricate behaviors with higher precision than conventional machine-learning methods \cite{17, 18}. These properties make such architectures a natural fit for BFRB detection, where subtle changes in hand proximity and thermal contact must be tracked over time.

Despite this progress, existing wearable BFRB studies have largely relied on single-modality or motion-plus-physiology sensing and have rarely fused fine-grained thermal and spatial-proximity cues within a deep spatio-temporal model; consequently, the objective, fine-grained discrimination of individual, anatomically similar BFRB gestures together with interpretable evidence of what drives model decisions remains underexplored. To address this gap, the Child Mind Institute developed the Helios wrist-worn device, which augments a standard IMU with five thermopile and five ToF sensors \cite{19}.

In this study, we propose a multimodal deep learning framework that fuses synchronized IMU, THM, and ToF streams to bridge kinematic motion and spatial context in behavioral monitoring. We then evaluate this framework on the Child Mind Institute dataset \cite{19}. Built on a hybrid CNN-GRU architecture with modality-specific autoencoders, the framework jointly models temporal and spatial dynamics while remaining sensitive to the subtle, episodic nature of these behaviors, distinguishing pathological BFRBs from routine daily actions. The contributions of this study are listed below:
\begin{itemize}
    \item \textbf{Multimodal Sensor Fusion:} We introduce a framework that integrates IMU, THM, and ToF data streams, and evaluate it to assess its improved accuracy over unimodal baselines.

    \item \textbf{Hybrid Spatio-Temporal Deep Learning Architecture:} We develop a hybrid CNN-GRU architecture that captures both spatial and temporal characteristics of wearable sensor data.

    \item \textbf{Autoencoder-Based Self-Supervised Pretraining:} We incorporate modality-specific autoencoder pretraining to initialize the dynamic sensor encoders using reconstruction-informed representations.

    \item \textbf{Explainability in Behavioral Sensing:} We implement post-hoc interpretability analysis using SHapley Additive exPlanations (SHAP) attributions and hierarchical clustering to reveal the relative importance of different sensor modalities and identify anatomical patterns in gesture misclassification.
\end{itemize}

The remainder of this paper is organized as follows. Section \ref{Section II} details the dataset, per-modality preprocessing, and the proposed architecture. Section \ref{Section III} presents binary and multi-class results with post-hoc interpretability and misclassification analyses. Section \ref{Section IV} discusses clinical implications, limitations, and future directions.

\section{Methods} \label{Section II}
\subsection{Dataset}
This study utilizes the “CMI - Detect Behavior with Sensor Data” dataset, provided by the Child Mind Institute \cite{19}. The dataset comprises sensor recordings from participants performing BFRBs and non-BFRB gestures while wearing the Helios wrist-worn device.
Data were collected in a controlled study where participants wore the Helios device on the wrist of their dominant arm \cite{19}. The device integrates three sensor types: (1) a CEVA Technologies, Inc. Inertial Measurement Unit (IMU; models BNO080/BNO085), measuring linear acceleration along three axes ($acc_x, acc_y, acc_z$ in m/s²) and providing orientation data ($rot_w, rot_x, rot_y, rot_z$) by fusing accelerometer, gyroscope, and magnetometer measurements; (2) five Melexis thermopile sensors (model MLX90632), capturing non-contact temperature in degrees Celsius ($thm_1$ to $thm_5$); and (3) five STMicroelectronics time-of-flight sensors (model VL53L7CX), recording proximity data across an 8$\times$8 grid per sensor (${t}_{1}^{v_0}$ to ${t}_{5}^{v_{63}}$), with values ranging from 0 to 254 (uncalibrated sensor units) or -1 for no response. Intermittent sensor communication issues resulted in missing values for a subset of thermopile and time-of-flight measurements in certain sequences.

The dataset contains a total of 8{,}151 unique sequences. These data were collected from 81 unique participants ($N=81$). To ensure a robust representation of behavioral variability, each participant provided an average of $\sim$7{,}098 replicates (min: 4{,}008; max: 10{,}848). Individual sequences have an average length of 70.54 time steps (min: 29; max: 700). Each sequence represents a trial with three phases: (1) a transition from a rest position to the target location; (2) a brief pause; and (3) execution of a gesture. Participants performed 18 distinct gestures, including 8 BFRB-like gestures (e.g., pulling hair above the ear) and 10 non-BFRB-like gestures (e.g., drinking from a cup), conducted in at least one of four body positions: sitting, sitting leaning forward with the non-dominant arm resting on the leg, lying on the back, or lying on the side. These gestures are listed in Table \ref{tab1}, with detailed descriptions and video examples provided in the dataset documentation \cite{19}. Sequences are labeled with sequence type (Target or Non-Target), gesture, orientation, behavior phase, sequence counter, and subject identifier \cite{19}.

The dataset includes participant demographics and anthropometric data, such as age (in years), sex (0: female; 1: male), handedness (0: left-handed; 1: right-handed), adult/child status (0: under 18 years; 1: adult), height (in cm), shoulder-to-wrist distance (in cm), and elbow-to-wrist distance (in cm). Participants were recruited through the Healthy Brain Network \cite{23}, a Child Mind Institute initiative \cite{19}.

\begin{table}[h]
	\caption{List of gestures included in the dataset.}
	\centering
	\begin{tabular}{ll}
		\toprule
		BFRB-Like Gestures (Target) & non-BFRB-Like Gestures (Non-Target) \\
		\midrule
		Above ear -- Pull hair      & Drink from bottle/cup \\
		Forehead -- Pull hairline   & Glasses on/off \\
		Forehead -- Scratch         & Pull air toward your face \\
		Eyebrow -- Pull hair        & Pinch knee/leg skin \\
		Eyelash -- Pull hair        & Scratch knee/leg skin \\
		Neck -- Pinch skin          & Write name on leg \\
		Neck -- Scratch             & Text on phone \\
		Cheek -- Pinch skin         & Feel around in tray and pull out an object \\
		                            & Write name in air \\
		                            & Wave hello \\
		\bottomrule
	\end{tabular}
	\label{tab1}
\end{table}

\subsection{Preprocessing}
All modalities were standardized to a fixed temporal extent of $T=128$ frames per sequence. This threshold was empirically selected because it fully preserves 95.13\% of all recorded sequences, thereby capturing complete gesture trajectories while mitigating the computational overhead of extreme outliers. Sequences shorter than $T$ were padded, and longer sequences were truncated. Padding respected modality semantics, with zeros used for IMU and THM channels, while $-1$ was reserved for ToF frames to denote ``no sensor return.'' A Boolean mask $\mathbf{M}\in\{0,1\}^{B\times T}$ encoded valid time steps and was propagated through recurrent packing operations as well as all masked reconstruction and classification losses, thereby ensuring that masked regions neither contributed to attention weighting nor influenced optimization. Here, $B$ denotes the batch size.

\subsubsection{IMU preprocessing}\label{imu-pre} Missing values within the sequence were imputed by linear interpolation to preserve local temporal trends. To resolve missing values at the sequence boundaries, we applied backward and forward filling using the nearest valid observations. From the tri-axial accelerations $(acc_x, acc_y, acc_z)$, we computed the \emph{acceleration magnitude} at each frame in Eq.~\eqref{eq:accmag}, as commonly applied in biomechanics~\cite{20}. 

\begin{equation}
    \label{eq:accmag}
    acc_{\text{mag}} = \sqrt{acc_x^2 + acc_y^2 + acc_z^2}
\end{equation}

Temporal derivatives were computed by first-order differences along time for each axis. The resulting \emph{jerk magnitude}, defined in Eq. \eqref{eq:jerkmag}, captures the rate of change of acceleration, a metric frequently used to quantify movement smoothness and coordination. This approach is informed by the minimum-jerk formulation prevalent in motor-control and biomechanics research, which posits that goal-directed human movements are optimized to minimize derivative fluctuations ~\cite{20,21}.

\begin{equation}
    \label{eq:jerkmag}
    \text{jerk}_{\text{mag}} = \sqrt{\big(\Delta acc_x\big)^2 + \big(\Delta acc_y\big)^2 + \big(\Delta acc_z\big)^2}
\end{equation}

where $\Delta$ represents the first-order temporal difference operator ($x_{t} - x_{t-1}$) applied to the tri-axial acceleration components.

To characterize orientation dynamics, we calculate the angular displacement between successive unit quaternions, $q_t$ and $q_{t-1}$. This relative rotation, or 'difference' quaternion ($\Delta q = q_t \otimes q_{t-1}^*$), is converted into a scalar \emph{angular-velocity proxy} ($\omega$) by extracting the rotation angle over a single sampling interval. Following standard quaternion kinematics, the angular change is derived from the scalar component of the unitized relative quaternion as shown in Eq. \eqref{eq:quatomega}~\cite{22}:
\begin{equation} \label{eq:quatomega} 
\omega = 2 \cdot \arccos(|\langle q_t, q_{t-1} \rangle|) 
\end{equation}
where $\langle q_t, q_{t-1} \rangle$ denotes the inner product of successive quaternions, providing a robust measure of angular change that is invariant to quaternion antipodal representation.

We extracted summary statistics for sequences and the corresponding newly created features, including mean, standard deviation, minimum, maximum, range, interquartile range (IQR), skewness, and kurtosis. We also created several other features: (i) zero-crossings count to capture sign changes; (ii) number of local peaks using the \emph{find\_peaks} algorithm from the SciPy signal processing library ~\cite{24}; (iii) dominant frequency obtained from the magnitude spectrum of the \emph{fast Fourier transform (FFT)} with unit sampling interval; and (iv) total spectral power as a proxy for overall energy in the frequency domain. This process resulted in a total of 156 static features for the IMU sensor.

\subsubsection{THM preprocessing}\label{thm-pre} To handle missing data, each thermopile channel first underwent linear interpolation across all gaps within the sequence. To guarantee complete sequences and impute any remaining missing values, particularly those at the extreme beginning or end of the recording, this interpolation was immediately followed by backward and forward filling. For every channel, we then computed sequence-level statistics such as mean, standard deviation, minimum, maximum, and range to quantify central tendency and dispersion. To capture distributional complexity beyond low-order moments, we constructed a 10-bin density-normalized histogram and computed its Shannon entropy, thereby quantifying irregularity and spread. Unlike statistical moments, which are exponentially skewed by the magnitude of extreme values, this histogram-based Shannon entropy evaluates probability mass, restricting an outlier's impact to a mere frequency count. In addition, we aggregated cross-channel behavior by taking the mean, standard deviation, and range across the channel means, which yields global THM descriptors that stabilize the representation in the presence of channel-specific noise and varying signal-to-noise ratios. This process resulted in a total of 33 static features for the THM sensor.

\subsubsection{ToF preprocessing} 
\begin{figure*}
    \centering
    \captionsetup[subfigure]{labelformat=empty}
    
    \begin{subfigure}[t]{\textwidth}
        \raggedright{\textbf{(A)}}\\[0.2em]
        \includegraphics[width=\linewidth]{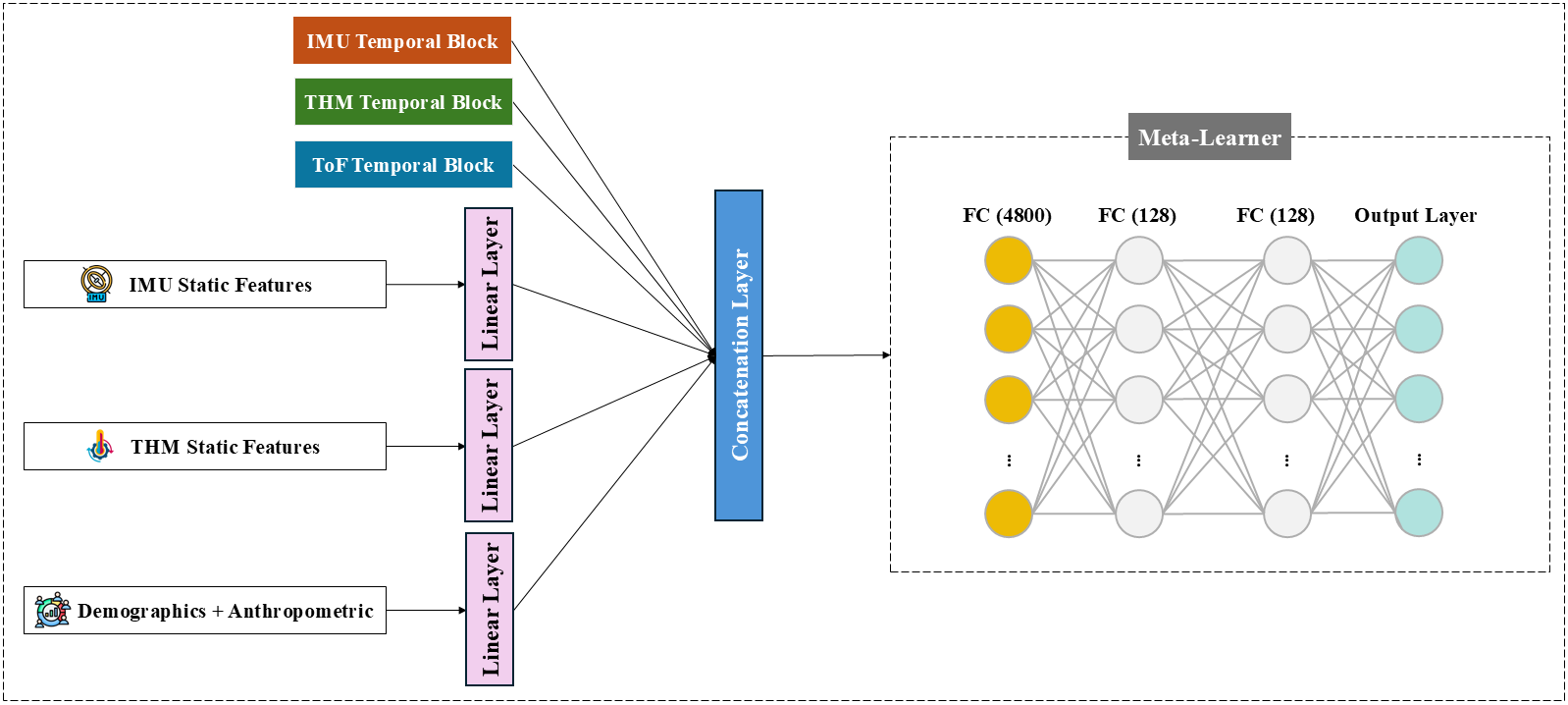}
        \phantomcaption
        \label{1a}
    \end{subfigure}

    \vspace{0.5em}

    \begin{subfigure}[t]{0.48\textwidth}
        \raggedright{\textbf{(B)}}\\[0.2em]
        \includegraphics[width=\linewidth, height=5cm]{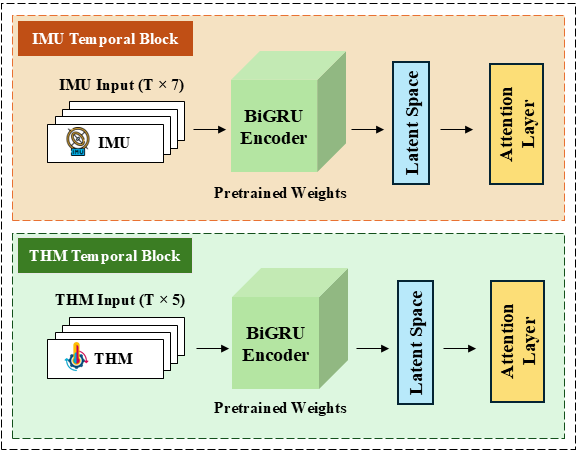}
        \phantomcaption
        \label{1b}
    \end{subfigure}\hfill
    \begin{subfigure}[t]{0.48\textwidth}
        \raggedright{\textbf{(C)}}\\[0.2em]
        \includegraphics[width=\linewidth, height=5cm]{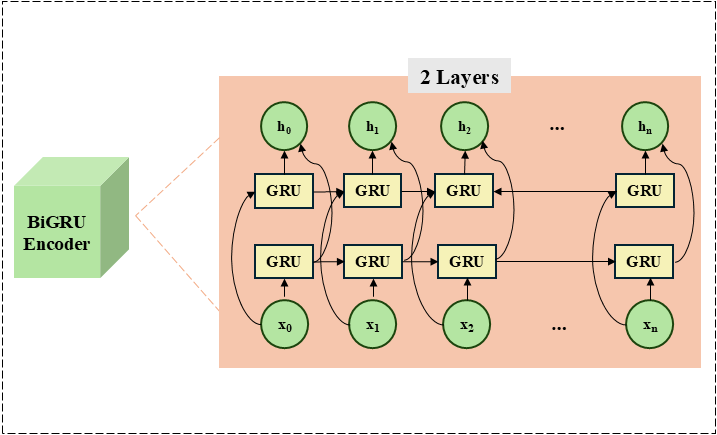}
        \phantomcaption
        \label{1c}
    \end{subfigure}

    \vspace{0.5em}

    \begin{subfigure}[t]{\textwidth}
        \raggedright{\textbf{(D)}}\\[0.2em]
        \includegraphics[width=\linewidth]{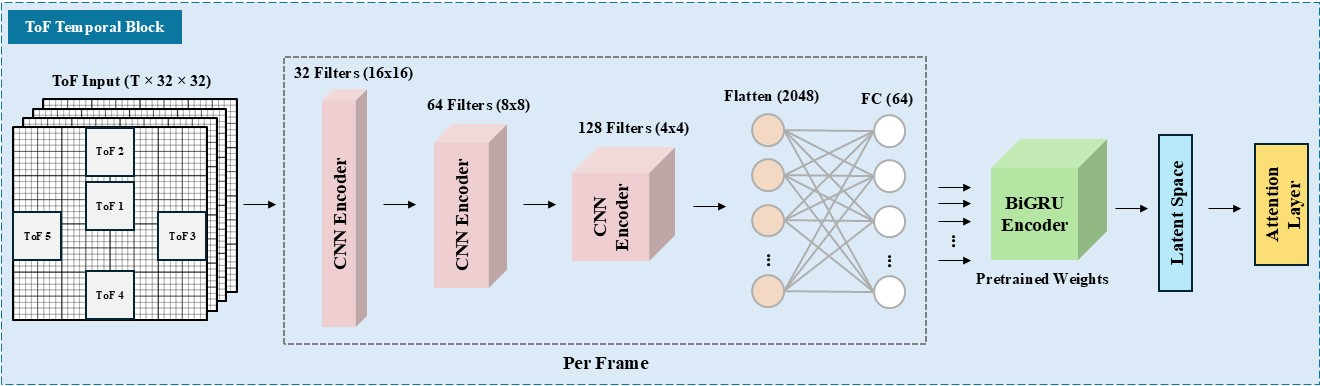}
        \phantomcaption
        \label{1d}
    \end{subfigure}

    \caption{Overview of the proposed multimodal fusion architecture. (A) Late-fusion Meta-Learner that combines dynamic sequence summaries and projected static embeddings for gesture classification. (B) IMU and THM processing pipeline using modality-specific pretrained Bidirectional GRU encoders and segment-wise attention. (C) Bidirectional GRU architecture showing the concatenation of forward and backward hidden states. (D) ToF processing pipeline, where CNN-extracted spatial embeddings are modeled by a pretrained Bidirectional GRU and summarized using segment-wise attention. Dynamic encoders in (B) and (D) are initialized from masked autoencoders pretrained using a self-supervised reconstruction objective and subsequently fine-tuned end-to-end.}
    \label{fig:1}
\end{figure*}

At each timestep, the ToF sensor provided up to five concurrent $8 \times 8$ proximity maps. To explicitly preserve the physical geometry of the sensor placement on the body, these patches were mapped onto a unified $32 \times 32$ spatial canvas based on their relative hardware locations (see Figure~\ref{1d}). Specifically, the top, inner-top, left, right, and bottom sensors (designated as sensors $2$, $1$, $5$, $3$, and $4$, respectively) were anchored at canvas origin coordinates (row, column) of $(0, 12)$, $(9, 12)$, $(14, 0)$, $(14, 24)$, and $(24, 12)$. This cross-like configuration resulted in a $1 \times 32 \times 32$ tensor per frame. Valid proximity values were normalized to a $[0, 1]$ scale. Empty spatial regions on the canvas, as well as frames with missing returns, were explicitly padded with a sentinel value of $-1$.

\subsection{Labeling}\label{label}
The model is trained with an 18-class output layer that matches the gesture set in the dataset (8 BFRBs and 10 non-BFRBs). For evaluation, we first define a binary mapping in which all eight BFRB-like gestures are treated as Target and the ten non-BFRB-like gestures are treated as Non-Target, consistent with Table \ref{tab1} and the dataset annotations. The binary classification score is obtained by summing the 18-class probabilities across all Target gestures, and the decision is positive when this aggregated Target probability meets or exceeds a threshold of 0.5. We also report a multi-class view that retains gesture specificity for the BFRB-like actions. For this purpose, a 9-class mapping is formed by keeping each Target gesture as its own class while merging all Non-Target gestures into a single Non-Target class.

\subsection{Proposed Methodology} \label{proposed_method}
The proposed methodology introduces a comprehensive multimodal fusion framework designed to detect and classify BFRBs. A schematic overview of the data flow and overall architecture is illustrated in Figure~\ref{fig:1}. The system operates by aggregating two sets of features: dynamic and static. Dynamic features include three synchronized sensor streams: kinematic movements from the IMU, thermal heat signatures from the THM, and spatial proximity maps from the ToF sensors. Static features comprise summary-statistic variables (see Sections \ref{imu-pre} and \ref{thm-pre}), demographics, and anthropometric data (see Figure~\ref{1a}).

We refer to the complete and integrated framework as the \textsc{ALL-Modality} model. Within this architecture, preprocessed sequences (i.e., IMU, THM, and ToF) with a fixed length of $T=128$ time steps are independently fed through dedicated, modality-specific temporal encoders (see Figures~\ref{1b} and ~\ref{1d}). These encoders extract high-level dynamic embeddings with a uniform hidden dimension of $H=64$. The embeddings are subsequently pooled into $S=12$ discrete segments via temporal attention. Concurrently, supplementary static context is established by integrating demographic and anthropometric data with static features from the IMU and THM signals (see Figure~\ref{1a}). The dynamic and static feature embeddings are then concatenated using a late-fusion strategy and passed to a late-fusion Meta-Learner, implemented as a multi-layer perceptron (MLP), to predict the gesture class.

We also proposed a self-supervised reconstruction-based pretraining strategy for initialization of encoders' weights. Specifically, the dynamic IMU, THM, and ToF streams are first modeled within modality-specific autoencoder architectures, where each encoder learns modality-specific temporal or spatio-temporal structure by supporting reconstruction of the original signal. The details of the autoencoder architectures, reconstruction objective, and encoder weight transfer procedure are described in the following sections.

\subsection{Autoencoders}
In this stage, we train encoder-decoder models to reconstruct the original input signals. The primary purpose of this phase is to obtain reconstruction-informed weights for the dynamic sensor encoders, which are later used to initialize the corresponding supervised encoders for IMU, THM, and ToF, separately. The decoder components are used only during pretraining and are discarded afterward.

\subsubsection{IMU/THM autoencoders}
For the IMU and THM sensors, the autoencoders utilize a sequence-to-sequence architecture based on Gated Recurrent Units (GRUs). To capture both past and future temporal context, the encoder consists of a two-layer bidirectional GRU with a hidden dimension of 64 nodes per direction, regularized with a 0.2 dropout rate. This bidirectional processing yields a 128-dimensional latent embedding for each valid timestep. To properly handle variable-length trials, the encoder operates on packed sequences defined by the Boolean validity mask, ensuring padded frames are explicitly ignored. The corresponding decoder is structured as a single-layer unidirectional GRU with a 64-node hidden state, followed by a linear projection layer that reconstructs the original multi-channel inputs (7 channels for IMU, 5 for THM). This sequence-to-sequence design preserves temporal granularity and encourages the encoder to retain information sufficient to reproduce signal dynamics, which aligns the latent space with what the downstream recognition encoders later exploit.

\subsubsection{ToF autoencoder}
For the ToF modality, a hybrid CNN-GRU autoencoder is utilized to model both spatial and temporal structures. The spatial encoder processes the $1 \times 32 \times 32$ ToF canvas through three consecutive convolutional blocks. Each block applies a $3 \times 3$ convolution, batch normalization, rectified linear activation (ReLU), $2 \times 2$ max-pooling, and a 0.2 dropout rate, with filter sizes progressively expanding from 32 to 64 to 128. The resulting $128 \times 4 \times 4$ feature map is flattened and linearly projected into a 64-dimensional spatial embedding per frame. This sequence of embeddings is then fed into a two-layer bidirectional GRU (64 hidden nodes per direction, 0.2 dropout), producing a 128-dimensional spatio-temporal latent representation for each valid timestep. The decoding pathway mirrors this architecture: a single-layer unidirectional GRU (64 hidden nodes) first reconstructs the temporal sequence. The spatial decoder then projects each timestep back to a $128 \times 4 \times 4$ tensor, applying three symmetrical stages of nearest-neighbor upsampling (scale factor 2) paired with $3 \times 3$ convolutions, batch normalization, and ReLU activations to progressively reconstruct the original $1 \times 32 \times 32$ spatial canvas. This pathway preserves frame-level locality while allowing the temporal module to capture motion-consistent factors in the latent sequence.

\subsubsection{Static autoencoders}
Demographic-anthropometric data and static IMU and THM features are handled during pretraining with independent, symmetric linear encoder-decoder branches. For each static branch, the encoder consists of a single fully connected linear layer that maps the respective input vector into a unified 64-dimensional latent space, immediately followed by 1D batch normalization. The corresponding decoder utilizes a single linear layer to project this 64-dimensional embedding back to the original input dimensionality for reconstruction. These static reconstruction branches allow the pretraining objective to incorporate the available non-sequential feature groups while maintaining a representation scale compatible with the 64-node static projections used during late fusion.

\subsubsection{Masked reconstruction loss}\label{Masked-MSE}
Reconstruction employs a masked mean-squared error (MSE) that is calculated exclusively over valid elements, ensuring that sequence padding does not influence optimization. For the ToF modality, the validity mask is applied across the entire frame tensor to explicitly ignore the $-1$ sentinels denoting missing returns or empty canvas space. The static branches (demographics and static features) utilize standard MSE. The total pretraining objective is formulated as the sum of all individual branch losses, and the best autoencoder configuration is selected based on the minimum validation reconstruction loss.

\subsubsection{Weight transfer}
After pretraining, the learned weights are copied into the corresponding modality-specific encoders of the supervised model. The decoder segments of autoendcoders are discarded and are not used during classification. All transferred weights are then unfrozen and fine-tuned end-to-end together with the segment-wise attention layers, static projection heads, and late-fusion Meta-Learner. Therefore, in the current implementation, autoencoder pretraining provides reconstruction-informed initialization for the dynamic encoders in the final supervised model. The supervised fine-tuning phase is regularized using the AdamW optimizer with a weight decay of $10^{-4}$, a network-wide dropout rate of $0.2$, and an early stopping mechanism based on validation performance.

\subsection{Encoders}
\subsubsection{Sequence encoders with segment-wise attention} \label{encoder1}
As illustrated in Figure \ref{1b}, each of the IMU and THM streams are modeled with independent, modality-specific two-layer bidirectional GRUs that do not share weights. These bidirectional GRUs, which are initialized with the parameters learned during the autoencoder pretraining phase, operate directly on the preprocessed sequence of length $T=128$ timesteps. Operating in batch-first mode with a hidden size of $H=64$ per direction, the encoder yields a latent sequence $\mathbf{Z} \in \mathbb{R}^{B \times T \times 2H}$. This sequence captures forward and backward temporal context while strictly ignoring padded positions via length-aware sequence packing. To obtain a fixed-length representation that preserves coarse temporal organization and achieves dimensionality reduction, the $T$-length timeline is partitioned along the time-index axis into $S=12$ contiguous segments. These segments are of roughly equal length in terms of time indices ($\lfloor T/S \rfloor$ frames per segment, with the final segment absorbing any remainder). Within each segment, a learned linear scoring function assigns one scalar score $e_t$ per valid timestep $t$, formulated as $e_t = \mathbf{w}^\top \mathbf{z}_t + b$, where $\mathbf{w} \in \mathbb{R}^{2H}$ and $b \in \mathbb{R}$ are the learnable weight and bias of the attention layer, respectively. The scores are then normalized over the valid timesteps within each segment using a masked softmax:
\begin{equation}
    \alpha_t =
    \frac{\exp(e_t)}
    {\sum_{j \in \mathcal{V}_s} \exp(e_j)},
    \quad t \in \mathcal{V}_s,
    \label{eq:masked_attention}
\end{equation}
where $\mathcal{V}_s$ denotes the set of valid timesteps within segment $s$, and $\alpha_t$ represents the normalized attention weight assigned to timestep $t$. Invalid or padded timesteps are assigned zero attention weight. These attention weights are then used to compute a weighted sum of the latent states $\mathbf{Z}$ over that segment.

The segment-level representation is then computed as the weighted sum of the valid latent states:
\begin{equation}
    \mathbf{z}_s =
    \sum_{t \in \mathcal{V}_s} \alpha_t \mathbf{z}_t,
    \label{eq:segment_representation}
\end{equation}
where $\mathbf{z}_s \in \mathbb{R}^{2H}$ denotes the attention-pooled representation of segment $s$.

A dropout rate of 0.2 is applied to each segment-level representation $\mathbf{z}_s$ after attention pooling for regularization. Finally, the $S$ segment summaries, each of size $2H$, are concatenated to form a single flattened descriptor of size $2H \times S$ (i.e., 1,536 dimensions) for each modality. The resulting flattened tensor is then fed into the late-fusion concatenation layer.

\subsubsection{Spatio-temporal ToF encoder}
As depicted in Figure \ref{1d}, the ToF modality is processed by a hybrid convolutional-recurrent pathway designed to handle the sequence of $1 \times 32 \times 32$ spatial proximity frames. First, each frame is embedded into a spatial feature vector using a three-block CNN. Each convolutional block consists of a $3 \times 3$ kernel, batch normalization, ReLU activation, $2 \times 2$ max-pooling, and a dropout regularization with a dropout rate of 0.2.
The network progressively expands the channel dimension from 1 to 32, 64, and 128 through learned convolutional filters, while max-pooling reduces the spatial resolution from 32×32 to 16×16, 8×8, and 4×4. Thus, the network trades spatial resolution for a richer set of learned feature maps. A subsequent flattening operation unrolls the $128 \times 4 \times 4$ tensor into a 2048-dimensional vector. A fully connected linear layer then projects this 2048-dimensional vector down to an $H$-dimensional (64-node) spatial embedding per frame. This sequence of per-frame embeddings is then passed through a two-layer bidirectional GRU identical to those used in the IMU/THM branches (see \ref{encoder1}), initializing with its respective pretrained autoencoder weights. The identical segment-wise attention mechanism ($S=12$) is then applied to the bidirectional GRU output to obtain a fixed-length ToF sequence summary. This hybrid design preserves local spatial structure within each frame, while the recurrent layer and attention mechanism summarize longer-range temporal dependencies and emphasize the most informative intervals of the gesture. Finally, the $S$ segment summaries are concatenated to yield a flattened descriptor of size $2H \times S$ (1,536 dimensions). This resulting tensor is finally fed into the late-fusion concatenation layer, where it is aggregated with the dynamic summaries of the other modalities and the static projections prior to classification.

\subsubsection{Static features}
demographic, anthropometric, and static IMU and THM features are incorporated through lightweight projection heads that normalize the raw data and place them on the same latent scale as the dynamic sequence summaries. Each static data stream employs one fully connected layer to project its respective input vector into an $H$-dimensional embedding (here, $H=64$). This projection is followed by a 1D batch normalization layer and a dropout layer with a dropout rate of 0.2. The only structural distinction among the branches is the choice of activation function: the static IMU and THM feature branches use ReLU activation, whereas the demographic and anthropometric branch uses the identity activation.

\subsection{Data Fusion}
All available modality embeddings are concatenated to form a single fused representation. For the full \textsc{ALL-Modality} configuration, this combined feature set integrates the three dynamic sequence summaries (IMU, THM, and ToF) and the three static projections (demographic and anthropometric data, static IMU features, and static THM features). Because each of the three dynamic sequence summaries has a size of $2H \times S$ and each of the three static projections has a size of $H$, the fused dimensionality mathematically equals $3H(2S+1)$. With $H=64$ and $S=12$, it yields a 4800-dimensional fused representation. This fused vector is then fed into an aggregation network (i.e., Meta-Learner), which comprises two fully connected hidden layers with 128 nodes each. Each hidden layer applies a linear transformation followed by 1D batch normalization, a ReLU activation function, and dropout with a rate of 0.2. The output layer produces 18 logits corresponding to the original gesture classes.

\subsection{Baseline Models}
To evaluate the added value of multimodal sensor fusion, we compared the proposed \textsc{ALL-Modality} model with three unimodal counterparts: IMU, THM, and ToF models. These unimodal models serve as ablation baselines, allowing us to quantify the relative contribution of each sensor stream under the same modeling framework. All unimodal baselines maintain the same underlying encoder, attention, and Meta-Learner design used in the proposed architecture, while isolating only one dynamic sensor stream at a time. For each unimodal configuration, the fused representation concatenates the sequence summary of the selected modality with the applicable static branches. Demographic and anthropometric features are included in all unimodal baselines, while static IMU and THM features are included only in their corresponding IMU and THM configurations, respectively. The ToF baseline includes the ToF sequence summary together with demographic and anthropometric features.

\subsection{Training and Validation Procedures}

We train the multimodal architecture in two sequential phases. First, a self-supervised pretraining phase initializes the modality-specific autoencoders to learn robust spatial and temporal representations. Next, a supervised fine-tuning phase integrates these pretrained encoders into the full classification pipeline for end-to-end optimization using labeled data. We employ a stratified, grouped train-validation-test split strategy for model evaluation. We used an independent held-out set $X^{test}$ ($n=1,627$ sequences from $16$ participants) as the test set. Stratification was performed on the gesture class labels to ensure balanced distributions across the splits, while grouping was applied strictly at the participant level to prevent participant overlap across subsets (i.e., avoid data leakage). The remaining data were split into $X^{train}$ ($n=5,810$ sequences from $58$ participants) and $X^{val}$ ($n=714$ sequences from $7$ participants) using the same strategy. Both autoencoder pretraining and supervised classifier training were performed on the training set, while the validation set was used for model selection and early stopping. The test set was reserved exclusively for final performance evaluation and post-hoc analysis.

\subsubsection{Pretraining Procedure}
During the unsupervised phase, the autoencoders are optimized using the AdamW optimizer with a learning rate of $\eta=0.001$. The objective function for pretraining is a masked MSE loss (see Section \ref{Masked-MSE}). The autoencoders are trained for up to 100 epochs. Model selection was determined using an early stopping mechanism based on the validation reconstruction loss, with a patience of 25 epochs. This patience value was adopted to reduce the likelihood of overly early termination of training and to accommodate transient fluctuations in validation loss that are common in high-frequency physiological and motion data.

\subsubsection{Supervised Training Procedure}
During the supervised fine-tuning phase, we train the entire pipeline using the AdamW optimizer ($\eta=0.001$). We use AdamW, which decouples weight decay from the adaptive gradient updates to improve generalization \cite{25}. To update the network's weights, we minimize the standard multi-class cross-entropy loss across all 18 original gesture classes. However, while cross-entropy drives the actual learning, we use a custom F1-based metric for model selection and stopping decision defined as the average of binary F1 (for Target vs. Non-Target problem) and macro-averaged F1 score (for the multi-class problem). We used F1 score instead of the training loss for validation to reduce the tendency of model to overfit to majority classes. We trained the model for 200 epochs with an early stopping criterion set for no improvement in 25 consecutive epochs.

\subsection{Evaluation and Metrics}\label{metrics}
Evaluation is conducted on the test set using the aggregated probability scores defined in Section \ref{label}. We report a \emph{binary} view that distinguishes Target from Non-Target with a fixed decision threshold of \(0.5\), and a \emph{9-class} view that preserves each Target gesture as its own class while merging all Non-Target gestures into one class. Let \(\mathrm{TP}\), \(\mathrm{FP}\), \(\mathrm{TN}\), and \(\mathrm{FN}\) denote, respectively, the counts of true positives, false positives, true negatives, and false negatives obtained from the confusion matrix for the class under consideration, either the positive class (here, Target) in the binary setting or a one-vs-rest (OVR) class in the 9-class setting. The core metrics are defined as follows.
\begin{equation}
    \mathrm{Precision}=\frac{\mathrm{TP}}{\mathrm{TP}+\mathrm{FP}}
\end{equation}
\begin{equation}
    \mathrm{Recall\ (Sensitivity)}=\frac{\mathrm{TP}}{\mathrm{TP}+\mathrm{FN}}
\end{equation}
\begin{equation}
    \mathrm{Specificity}=\frac{\mathrm{TN}}{\mathrm{TN}+\mathrm{FP}}
\end{equation}
\begin{equation}
    \mathrm{F1\ score}=\frac{2\,\mathrm{Precision}\cdot \mathrm{Recall}}{\mathrm{Precision}+\mathrm{Recall}}
\end{equation}

Receiver operating characteristic area under the curve (ROC-AUC) summarizes discrimination across classification thresholds. In the 9-class setting, metrics are computed per class (OVR approach) and then macro-averaged to weight classes equally. All computations operate on these aggregated probabilities and their correspondingly merged ground-truth labels to remain consistent with the training head and the mappings described in Section \ref{label}.

\subsection{Post-Hoc Analysis}
We conducted a two-part post-hoc analysis to examine the network's decision-making and error dynamics. First, we used SHAP to quantify the modality and feature-level contributions driving predictions. Second, we applied hierarchical clustering to misclassification patterns to identify overlapping behaviors and systematic confusion.

\subsubsection{SHAP explanations on fused representations}
Interpretability was assessed for the \textsc{ALL-Modality} model by explaining the fused classifier input, including the concatenation of IMU, THM, and ToF sequence summaries together with static projections (i.e., demographics, anthropometric data, static IMU features, static THM features). We used Deep SHAP (DeepExplainer) \cite{26} with a background set of fused vectors drawn from the training split and computed attributions with respect to the predicted 18-class logit.

Let $x \in \mathbb{R}^{F}$ denote the fused feature vector and $f_c(x)$ the logit for class $c \in \{1,\dots,C\}$. The predicted class is
\begin{equation}
    \hat{c}(x) = \arg\max_{c} f_c(x).
    \label{eq:pred-class}
\end{equation}

Deep SHAP yields per-feature attributions $\phi^{(c)}_j(x)$ for class $c$ and fused feature index $j \in \{1,\dots,F\}$. We select the attribution vector for the predicted class:
\begin{equation}
    \phi_j(x) = \phi^{(\hat{c}(x))}_j(x), \quad j=1,\dots,F.
    \label{eq:pred-class-shap}
\end{equation}

Let $\{G_k\}_{k=1}^{K}$ be a partition of disjoint feature indices into modality-aligned fusion groups (here, $K=6$: IMU sequence, THM sequence, ToF sequence, demographics + anthropometric, static IMU features, static THM features). The per-sample group contribution is the sum of absolute SHAP values within each group:
\begin{equation}
    s_k(x) = \sum_{j \in G_k} \big|\phi_j(x)\big|, \quad k=1,\dots,K.
    \label{eq:block-sum}
\end{equation}

For reporting, we use the average contribution magnitude across the test set $\{x_i\}_{i=1}^{N}$ formulated as:
\begin{equation}
    \bar{s}_k = \frac{1}{N}\sum_{i=1}^{N} \sum_{j \in G_k} \big|\phi_j(x_i)\big|, \quad k=1,\dots,K.
    \label{eq:block-mean}
\end{equation}

Since the group sizes differ, we also report the average contribution per feature, defined as:
\begin{equation}
    \bar{s}_k^{'} = \frac{\bar{s}_k}{|G_{k}|}, \quad k=1,\dots,K.
    \label{eq:block-mean2}
\end{equation}
where $|G_{k}|$ denotes the size of feature set $k$.

\subsubsection{Hierarchical Clustering of Misclassification Patterns}
To examine systematic misclassification patterns among the gesture classes, we derived a dendrogram from the grouped test set confusion matrix in the 9-class setting (rows correspond to ground truth and columns to predictions). The Non-Target class was excluded prior to normalization. We then row-normalized the remaining matrix to obtain conditional prediction distributions $M_{uv} = P(\hat{c}=v \mid c=u)$ with $\sum_{v} M_{uv} = 1$ for each class $u$. For each pair of classes $(u, v)$, we defined a symmetric score
\begin{equation}
    \gamma_{uv} = \gamma_{vu} = \tfrac{1}{2}\left(M_{uv} + M_{vu}\right),
\end{equation}

which captures how frequently the model misclassifies class $u$ as $v$ and vice versa. A large value of $\gamma_{uv}$ shows that the two classes are similar in characteristics, making it challenging for the model to distinguish between them. A corresponding distance matrix was then obtained through the relationship
\begin{equation}
    d_{uv} = 1 - \gamma_{uv},
\end{equation}

with $d_{uu} = 0, \: \forall \, u \in \{1, 2, \ldots, C\} \setminus \{c_{\text{Non-Target}}\}$. Hierarchical clustering with the average linkage method was then applied to construct the dendrogram.

\section{Results} \label{Section III}
This Section presents the performance of the proposed models, evaluating their ability to classify BFRB versus non-BFRB gestures with binary and multi-class scopes. We then discuss the findings from post-hoc interpretability analyses.

\subsection{Classification Results}
The multimodal model was evaluated on a held-out test set using standard classification metrics described in Section \ref{metrics}. Table~\ref{tab:combined_performance} shows the classification results for both binary and multi-class tasks across the full multimodal configuration (\textsc{ALL-Modality}) and the unimodal baselines (IMU, THM, and ToF).
The \textsc{ALL-Modality} model, integrating IMU, THM, and ToF data with static features, demographic, and anthropometric data, achieved the highest performance with an F1 score of 0.985, precision of 0.977, recall of 0.992, specificity of 0.960, and ROC-AUC of 0.997. This outperformed the unimodal baselines: IMU (F1 = 0.972), THM (F1 = 0.945), and ToF (F1 = 0.960). The binary classification task appears relatively less challenging, as all models achieved strong F1 scores. However, the \textsc{ALL-Modality} model still provided a slight performance advantage, suggesting that multimodal sensor fusion offers complementary information beyond any single sensor stream.

\begin{table*}[h]
	\caption{Classification performance for binary and multi-class tasks}
	\centering
	\begin{tabular}{llccccc}
		\toprule
		Task & Modality & Precision & Recall & Specificity & F1 score & ROC-AUC \\
		\midrule
		Binary      & \textsc{ALL-Modality} & \textbf{0.977} & \textbf{0.992} & \textbf{0.960} & \textbf{0.985} & \textbf{0.997} \\
		            & IMU                   & 0.967          & 0.978          & 0.944          & 0.972          & 0.995          \\
		            & THM                   & 0.960          & 0.931          & 0.934          & 0.945          & 0.979          \\
		            & ToF                   & 0.938          & 0.983          & 0.889          & 0.960          & 0.986          \\
		\midrule
		Multi-Class & \textsc{ALL-Modality} & \textbf{0.709} & \textbf{0.704} & \textbf{0.973} & \textbf{0.700} & \textbf{0.963} \\
		            & IMU                   & 0.487          & 0.487          & 0.953          & 0.484          & 0.894          \\
		            & THM                   & 0.485          & 0.477          & 0.951          & 0.470          & 0.887          \\
		            & ToF                   & 0.655          & 0.627          & 0.964          & 0.614          & 0.940          \\
		\bottomrule
	\end{tabular}
	\label{tab:combined_performance}
\end{table*}

For multi-class evaluation, the \textsc{ALL-Modality} model attained a macro F1 score of 0.700, precision of 0.709, recall of 0.704, specificity of 0.973, and ROC-AUC of 0.963, surpassing IMU (F1 = 0.484), THM (F1 = 0.470), and ToF (F1 = 0.614). Although the multi-class F1 score represents a decline from binary performance, this reduction was anticipated given the increased granularity required to differentiate among closely related BFRB gestures, such as those involving similar hand-to-face interactions. Nonetheless, the results remain promising, demonstrating the model's capability to handle fine-grained classification in a challenging domain where gestural similarities contribute to classification difficulties.

Per-class F1 scores for the \textsc{ALL-Modality} model are presented in Table~\ref{tab:perclass}, further illustrating this variability, with the Non-Target class achieving the highest F1 (0.97), indicative of robust separation from BFRBs.

\begin{table}[h]
	\caption{Class-specific F1 scores for the \textsc{ALL-Modality} model}
	\centering
	\begin{tabular}{lc}
		\toprule
		Class & F1 score \\
		\midrule
		Non-Target                  & 0.973 \\
		Above ear -- pull hair      & 0.877 \\
		Forehead -- scratch         & 0.746 \\
		Forehead -- pull hairline   & 0.734 \\
		Neck -- pinch skin          & 0.667 \\
		Cheek -- pinch skin         & 0.609 \\
		Neck -- scratch             & 0.604 \\
		Eyelash -- pull hair        & 0.594 \\
		Eyebrow -- pull hair        & 0.496 \\
		\bottomrule
	\end{tabular}
	\label{tab:perclass}
\end{table}

Among BFRBs, \emph{Above ear -- pull hair} (0.877) and \emph{Forehead -- scratch} (0.746) showed strong performance, while \emph{Eyelash -- pull hair} (0.594) and \emph{Eyebrow -- pull hair} (0.496) exhibited lower scores, likely due to overlapping kinematic and proximity patterns. To better understand the anatomical and behavioral overlaps driving these specific classification bottlenecks, we explore these systematic misclassification patterns further in the post-hoc hierarchical clustering analysis detailed in the following Section.

\subsection{Post-Hoc Analyses}
To assess the relative contributions of input features to model predictions, SHAP values were computed on a test set. Table~\ref{tab:shap} ranks sensor groups by the sum of mean absolute SHAP values ($\bar{s}_{k}$), alongside the per-feature contribution ($\bar{s}_{k}^{'}$), which normalizes by group size. ToF sequences exhibited the highest total contribution (12.72), followed by IMU sequences (10.34), THM sequences (1.68), THM static features (0.79), IMU static features (0.73), and demographics (0.36). These rankings reveal that temporal sequences from the ToF and IMU sensors exert the most substantial overall impact on the prediction, likely due to their ability to effectively capture spatial proximity and dynamic gestural patterns over time. On a per-feature basis, however, the compact static descriptors are the most informative, with THM (0.01241) and IMU (0.01135) static features ranking above the ToF sequences (0.00828); this indicates that their large total attribution is distributed across many dimensions, whereas a few static features each carry disproportionate weight. Notably, ToF remains a top contributor under both measures, reinforcing the central role of spatial-proximity information.

\begin{table}[h]
	\caption{Aggregated SHAP values across feature subsets, ranked by total contribution $\bar{s}_{k}$.}
	\centering
	\begin{tabular}{clcc}
		\toprule
		Rank & Feature Subset & $\bar{s}_{k}$ & $\bar{s}_{k}^{'}$ \\
		\midrule
		1 & ToF sequences       & 12.72 & 0.00828 \\
		2 & IMU sequences       & 10.34 & 0.00673 \\
		3 & THM sequences       & 1.68  & 0.00109 \\
		4 & THM static features & 0.79  & 0.01241 \\
		5 & IMU static features & 0.73  & 0.01135 \\
		6 & Demographics        & 0.36  & 0.00569 \\
		\bottomrule
	\end{tabular}
	\label{tab:shap}
\end{table}

\begin{figure}[h]
	\centering
	\includegraphics[width=0.55\linewidth]{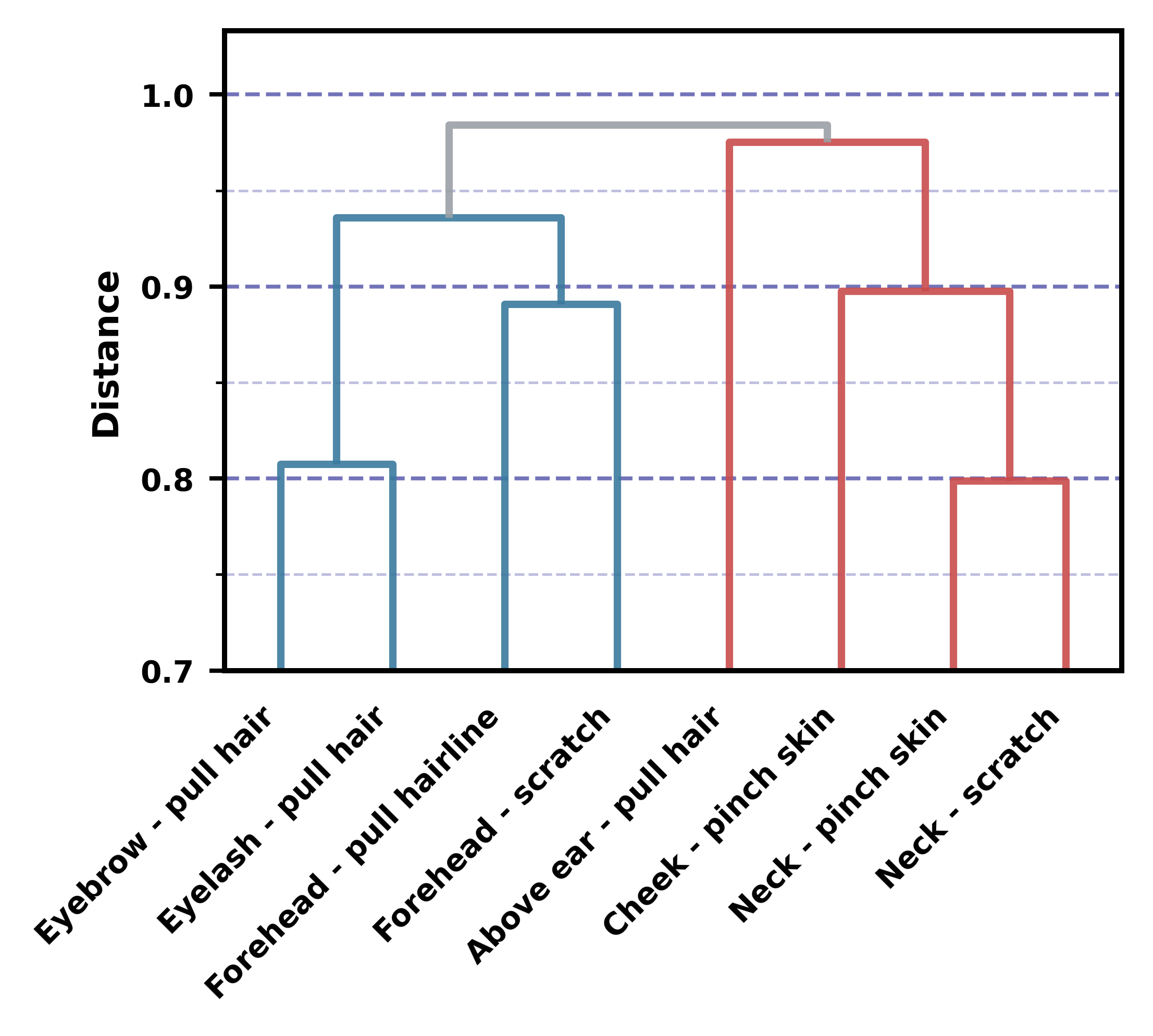}
	\caption{Hierarchical clustering of the misclassification patterns from reported classes (excluding the Non-Target category). The y-axis is cropped to improve visual clarity.}
	\label{fig:dendrogram}
\end{figure}

The dendrogram in Figure~\ref{fig:dendrogram} offers a hierarchical view of gesture-level misclassification patterns among the eight BFRBs. It summarizes how often classes are mistaken for one another; shorter branches indicate greater overlap in the model’s error patterns. Starting from the lowest linkages, two distinct pairs form almost simultaneously at a distance of approximately 0.80: the eye-region pair (\emph{Eyebrow – pull hair} and \emph{Eyelash – pull hair}) and the neck-region pair (\emph{Neck – pinch skin} and \emph{Neck – scratch}). Slightly higher in the hierarchy, \emph{Forehead – pull hairline} and \emph{Forehead – scratch} merge into a forehead pair. As the clustering progresses, the tree clearly divides into two broader branches based on anatomical proximity. On the left, the eye pair merges directly with the forehead pair to form an upper-face cluster. On the right, \emph{Cheek – pinch skin} merges with the neck pair to form a lower/side cluster. Finally, \emph{Above ear – pull hair} remains comparatively distinct, only merging with the lower-face/neck branch near the top of the hierarchy. Overall, the dendrogram indicates strong within-region misclassifications (eye with eye; forehead with forehead; neck with neck) and comparatively greater distinctiveness for the ear-region gesture.

Taken together, the dendrogram structure and the per-class F1 scores are mutually consistent. The neck pair and the eyebrow--eyelash pair sit at the lowest linkage heights, reflecting strong mutual misclassification and matching their lower respective F1 scores (0.50--0.67). By contrast, the forehead pair merges at a higher distance, indicating a lower misclassification rate; this aligns with their higher F1 scores (0.73--0.75) and suggests these classes are more separable despite their shared anatomical region. \emph{Above ear – pull hair} remains isolated until late in the hierarchy, which perfectly corroborates its strong F1 score (0.88). Overall, classes that cluster early tend to be harder to distinguish, while those that merge later achieve better per-class F1 performance.

A coherent picture emerges indicating that misclassifications are driven more by contact location. The dendrogram consistently merges clusters within the same or adjacent anatomical regions (e.g., eye with forehead; cheek with neck), and the lowest per-class F1 scores appear within these tightly clustered groups.

\section{Conclusion and Future Work} \label{Section IV}
This study presents a multimodal deep learning framework for the reliable detection and fine-grained characterization of BFRBs using wrist-worn wearable sensors. Leveraging IMU, THM, and ToF modalities from the Child Mind Institute’s Helios device, the proposed architecture processes each sensor stream through dedicated encoders, for each the parameters are initialized from a pretrained autoencoder model. These representations are then pooled using temporal segment-wise attention to create fixed-length dynamic summaries. Finally, these summaries are concatenated with static demographic, anthropometric, and static features in a late-fusion layer, passing through a Meta-Learner to predict the gesture class. Evaluation on a subject-grouped held-out test set demonstrates that this multimodal fusion enhances discriminative performance, achieving near-perfect binary detection and strong multi-class recognition despite the inherent similarities among BFRB gestures.

Beyond high classification performance, the incorporation of post-hoc interpretability methods provides transparent insights into the model's reasoning and gesture-level confusion patterns. The SHAP-based feature attribution analysis explicitly confirmed that the network relies heavily on the complementary value of ToF and IMU signals to capture spatial proximity and dynamic motion. Furthermore, the hierarchical clustering analysis revealed a clear anatomical structure in the model’s errors, underscoring the practical challenges of distinguishing spatially adjacent BFRBs. 

In practice, this framework offers substantial benefits for both clinical professionals and individuals affected by BFRBs. For clinicians, the ability to objectively and continuously monitor BFRB episodes outside of clinical environments addresses the subjectivity and recall biases of traditional self-reporting. This provides actionable, data-driven insights into behavioral triggers and episode frequencies. For patients, integrating this lightweight, wearable-based detection system into digital mental health applications could facilitate real-time awareness and prompt timely, personalized behavioral interventions.

Despite these strong results, several limitations must be acknowledged to guide future research. The current evaluation relies on data collected in controlled settings, which may not fully encapsulate the variability of BFRBs in unconstrained, daily-living environments. Additionally, our hierarchical clustering indicated that misclassifications are primarily driven by contact location. These findings suggest targeted improvements such as adding region-aware priors, refining ToF spatial calibration, and explicitly modeling hand-face geometry to reduce errors among anatomically adjacent gestures. Future work should also explore in-the-wild deployments to validate model robustness and investigate dynamic segmentation techniques for continuous streaming data. Finally, incorporating additional physiological signals could support closed-loop adaptive interventions by linking changes in emotional or stress-related states with the occurrence of BFRB-related motor behaviors.

\section*{Acknowledgment}
The authors acknowledge the use of Google's Gemini for language refinement throughout the manuscript. The authors fully reviewed the content and take responsibility for the final work.

\bibliographystyle{unsrtnat}
\bibliography{CMI}  

@article{1,
  title={Prevalence, gender correlates, and co-morbidity of trichotillomania},
  author={Grant, Jon E and Dougherty, Darin D and Chamberlain, Samuel R},
  journal={Psychiatry research},
  volume={288},
  pages={112948},
  year={2020},
  publisher={Elsevier}
}

@article{2,
  title={Prevalence of skin picking (excoriation) disorder},
  author={Grant, Jon E and Chamberlain, Samuel R},
  journal={Journal of psychiatric research},
  volume={130},
  pages={57--60},
  year={2020},
  publisher={Elsevier}
}

@article{3,
  title={Prevalence of body-focused repetitive behaviors in a diverse population sample--rates across age, gender, race and education},
  author={Moritz, Steffen and Scheunemann, Jakob and Jelinek, Lena and Penney, Danielle and Schmotz, Stella and Hoyer, Luca and Grudzie{\'n}, Dominik and Aleksandrowicz, Adrianna},
  journal={Psychological Medicine},
  volume={54},
  number={8},
  pages={1552--1558},
  year={2024},
  publisher={Cambridge University Press}
}

@article{4,
  title={Trichotillomania and skin-picking disorder: an update},
  author={Grant, Jon E and Chamberlain, Samuel R},
  journal={Focus},
  volume={19},
  number={4},
  pages={405--412},
  year={2021},
  publisher={American Psychiatric Association Washington, DC}
}

@article{5,
  title={Feeling uncomfortable in your own skin: a qualitative study of problematic skin picking in Italian women},
  author={Montali, Lorenzo and Garnieri, Sara},
  journal={Current Psychology},
  volume={43},
  number={14},
  pages={12870--12881},
  year={2024},
  publisher={Springer}
}

@article{6,
  title={A wearable artificial intelligence feedback tool (wrist angel) for treatment and research of obsessive compulsive disorder: protocol for a nonrandomized pilot study},
  author={L{\o}nfeldt, Nicole Nadine and Clemmensen, Line Katrine Harder and Pagsberg, Anne Katrine},
  journal={JMIR research protocols},
  volume={12},
  number={1},
  pages={e45123},
  year={2023},
  publisher={JMIR Publications Inc., Toronto, Canada}
}

@article{7,
  title={Wearable, environmental, and smartphone-based passive sensing for mental health monitoring},
  author={Sheikh, Mahsa and Qassem, Meha and Kyriacou, Panicos A},
  journal={Frontiers in digital health},
  volume={3},
  pages={662811},
  year={2021},
  publisher={Frontiers Media SA}
}

@article{8,
  title={Anxiety and body-focused repetitive behaviors: A systematic review and meta-analysis of comorbidity rates and symptom associations},
  author={Barber, Kathryn E and Cram, Isabella F and Smith, Elyse C and Capel, Leila K and Snorrason, Ivar and Woods, Douglas W},
  journal={Journal of psychiatric research},
  volume={181},
  pages={80--90},
  year={2025},
  publisher={Elsevier}
}

@article{9,
  title={Heart rate variability in obsessive compulsive disorder in comparison to healthy controls and as predictor of treatment response},
  author={Olbrich, Hanife and Jahn, Ina and Stengler, Katarina and Seifritz, Erich and Colla, Michael},
  journal={Clinical Neurophysiology},
  volume={138},
  pages={123--131},
  year={2022},
  publisher={Elsevier}
}

@inproceedings{10,
  title={Anticipatory detection of compulsive body-focused repetitive behaviors with wearables},
  author={Searle, Benjamin Lucas and Spathis, Dimitris and Constantinides, Marios and Quercia, Daniele and Mascolo, Cecilia},
  booktitle={Proceedings of the 23rd International Conference on Mobile Human-Computer Interaction},
  pages={1--15},
  year={2021}
}

@article{11,
  title={Thermal sensors improve wrist-worn position tracking},
  author={Son, Jake J and Clucas, Jon C and White, Curt and Krishnakumar, Anirudh and Vogelstein, Joshua T and Milham, Michael P and Klein, Arno},
  journal={NPJ digital medicine},
  volume={2},
  number={1},
  pages={15},
  year={2019},
  publisher={Nature Publishing Group UK London}
}

@article{12,
  title={Robust Multimodal Learning Framework for Intake Gesture Detection Using Contactless Radar and Wearable IMU Sensors},
  author={Wang, Chunzhuo and Hallez, Hans and Vanrumste, Bart},
  journal={IEEE Journal of Biomedical and Health Informatics},
  year={2026},
  publisher={IEEE}
}

@article{13,
  title={Optimizing Mental Stress Detection via Heart Rate Variability Feature Selection},
  author={Behradfar, Mohsen and Roy, Shotabdi and Nuamah, Joseph},
  journal={Sensors},
  volume={25},
  number={13},
  pages={4154},
  year={2025},
  publisher={MDPI}
}

@article{14,
  title={Machine learning for multimodal mental health detection: a systematic review of passive sensing approaches},
  author={Khoo, Lin Sze and Lim, Mei Kuan and Chong, Chun Yong and McNaney, Roisin},
  journal={Sensors},
  volume={24},
  number={2},
  pages={348},
  year={2024},
  publisher={MDPI}
}

@article{15,
  title={Deep learning--based multimodal data fusion: Case study in food intake episodes detection using wearable sensors},
  author={Bahador, Nooshin and Ferreira, Denzil and Tamminen, Satu and Kortelainen, Jukka and others},
  journal={JMIR mHealth and uHealth},
  volume={9},
  number={1},
  pages={e21926},
  year={2021},
  publisher={JMIR Publications Inc., Toronto, Canada}
}

@article{16,
  title={Fatigue monitoring using wearables and AI: Trends, challenges, and future opportunities},
  author={Kakhi, Kourosh and Jagatheesaperumal, Senthil Kumar and Khosravi, Abbas and Alizadehsani, Roohallah and Acharya, U Rajendra},
  journal={Computers in Biology and Medicine},
  volume={195},
  pages={110461},
  year={2025},
  publisher={Elsevier}
}

@article{17,
  title={Deep learning in human activity recognition with wearable sensors: A review on advances},
  author={Zhang, Shibo and Li, Yaxuan and Zhang, Shen and Shahabi, Farzad and Xia, Stephen and Deng, Yu and Alshurafa, Nabil},
  journal={Sensors},
  volume={22},
  number={4},
  pages={1476},
  year={2022},
  publisher={MDPI}
}

@article{18,
  title={Recognition of sports and daily activities through deep learning and convolutional block attention},
  author={Mekruksavanich, Sakorn and Phaphan, Wikanda and Hnoohom, Narit and Jitpattanakul, Anuchit},
  journal={PeerJ Computer Science},
  volume={10},
  pages={e2100},
  year={2024},
  publisher={PeerJ Inc.}
}

@misc{19,
    author = {Laura Newman and David LoBue and Arianna Zuanazzi and
              Florian Rupprecht and Luke Mears and Roxanne McAdams and
              Erin Brown and Yanyi Wang and Camilla Strauss and Arno Klein and
              Lauren Hendrix and Maki Koyama and Josh To and Curt White and
              Yuki Kotani and Michelle Freund and Michael Milham and
              Gregory Kiar and Martyna Plomecka and Sohier Dane and
              Maggie Demkin},
    title = {{CMI} -- Detect Behavior with Sensor Data},
    year = {2025},
    howpublished = {\url{https://kaggle.com/competitions/cmi-detect-behavior-with-sensor-data}},
    note = {Kaggle competition}
}

@book{20,
  title={Biomechanics and motor control of human movement},
  author={Winter, David A},
  year={2009},
  publisher={John wiley \& sons}
}

@article{21,
  title={The coordination of arm movements: an experimentally confirmed mathematical model},
  author={Flash, Tamar and Hogan, Neville},
  journal={Journal of neuroscience},
  volume={5},
  number={7},
  pages={1688--1703},
  year={1985},
  publisher={Society for Neuroscience}
}

@book{22,
  title={Quaternions and rotation sequences: a primer with applications to orbits, aerospace, and virtual reality},
  author={Kuipers, Jack B},
  year={1999},
  publisher={Princeton university press}
}

@misc{23,
  author = {{Child Mind Institute}},
  title = {Healthy Brain Network},
  year = {2026},
  howpublished = {\url{https://childmind.org/science/global-open-science/healthy-brain-network/}},
  note = {Accessed: 2026-02-02}
}

@misc{24,
  title        = {find\_peaks --- SciPy v1.17.0 Manual},
  howpublished = {\url{https://docs.scipy.org/doc/scipy/reference/generated/scipy.signal.find_peaks.html}},
  note         = {Accessed: 2026-04-03}
}

@article{25,
  title={Decoupled weight decay regularization},
  author={Loshchilov, Ilya and Hutter, Frank},
  journal={arXiv preprint arXiv:1711.05101},
  year={2017}
}

@article{26,
  title={A unified approach to interpreting model predictions},
  author={Lundberg, Scott M and Lee, Su-In},
  journal={Advances in neural information processing systems},
  volume={30},
  year={2017}
}

@inproceedings{27,
  author    = {Mohsen Behradfar and Samaneh Rezaeimanesh and Mohammad Fili and Guiping Hu},
  title     = {Multimodal Wearable Sensor Fusion for Repetitive Behavior Detection},
  booktitle = {Proceedings of the IISE Annual Conference \& Expo},
  address   = {Arlington, TX, USA},
  year      = {2026}
}






\end{document}